\documentclass[journal]{IEEEtran}
\usepackage{graphicx}
\usepackage{cite}
\usepackage{amssymb}
\usepackage{amsmath}
\usepackage{enumerate}
\usepackage{booktabs}
\usepackage[ruled]{algorithm2e}
\newcommand{\algcomment}[1]{\textcolor{gray}{\texttt{\# #1}}}
\usepackage{color}
\usepackage{colortbl}
\usepackage[colorlinks,linkcolor=red,anchorcolor=blue,citecolor=green]{hyperref}
\usepackage{bbm}
\usepackage{array}
\usepackage{multirow}
\usepackage[table,xcdraw]{xcolor}

\usepackage[caption=false,font=footnotesize]{subfig}
\begin{document}
%
\title{Toward Realistic Remote Sensing Dataset Distillation with Discriminative Prototype-guided Diffusion}
%
%
%

\author{Yonghao~Xu,~\IEEEmembership{Member,~IEEE,}
        ~Pedram~Ghamisi,~\IEEEmembership{Senior Member,~IEEE,}
        and~Qihao~Weng,~\IEEEmembership{Fellow,~IEEE}

\thanks{This work was supported in part by the Swedish Research Council with grant agreement no. 2024-05652, and the Wallenberg Artificial Intelligence, Autonomous Systems and Software Program (WASP), funded by Knut and Alice Wallenberg Foundation. The computational resources were provided by the National Academic Infrastructure for Supercomputing in Sweden (NAISS) at C3SE, partially funded by the Swedish Research Council through grant agreement no. 2022-06725, and by the Berzelius resource, provided by the Knut and Alice Wallenberg Foundation at the National Supercomputer Centre.}
\thanks{Y. Xu is with the Computer Vision and Learning Systems, Department of Electrical Engineering, Linköping University, 581 83 Linköping, Sweden (e-mail: yonghao.xu@liu.se).}
\thanks{P. Ghamisi is with Helmholtz-Zentrum Dresden-Rossendorf, Responsible AI Group, 09599 Freiberg, Germany and also with the University of Iceland, Faculty of Electrical and Computer Engineering, Reykjavik 102, Iceland (e-mail: p.ghamisi@hzdr.de).}
\thanks{Q. Weng is with JC STEM Lab of Earth Observations, Department of Land Surveying and Geo-Informatics, the Research Centre for Artificial Intelligence in Geomatics, and the Research Institute for Land and Space, The Hong Kong Polytechnic University, Hung Hom, Kowloon, Hong Kong (e-mail: qihao.weng@polyu.edu.hk).}
}

%
%

\markboth{Accepted by IEEE Transactions on Geoscience and Remote Sensing}%
{Shell \MakeLowercase{et al.}: Bare Demo of IEEEtran.cls for IEEE Journals}
%



\maketitle

\begin{abstract}
Recent years have witnessed the remarkable success of deep learning in remote sensing image interpretation, driven by the availability of large-scale benchmark datasets. However, this reliance on massive training data also brings substantial storage and computational costs. To address this challenge, this study introduces the concept of dataset distillation into the field of remote sensing image interpretation for the first time. Specifically, we propose discriminative prototype-guided diffusion (DPD), a diffusion-based generative distillation framework that condenses a large-scale remote sensing dataset into a compact and representative distilled dataset. To improve the semantic fidelity and diversity of the synthesized samples, we extract representative prototypes for each category in the latent space. We then construct hyperspherical semantic anchors around the prototypes to guide the reverse denoising trajectory. Furthermore, to enhance the discriminative quality of the generated samples, multiple candidates are generated for each prototype and ranked by a latent classifier using a logit-margin criterion, with the most discriminative candidates selected to form the final distilled dataset. Experiments on three high-resolution remote sensing scene classification benchmarks show that the proposed method can distill realistic, diverse, and discriminative samples for downstream model training. Code and pre-trained models are available online (https://github.com/YonghaoXu/DPD).
\end{abstract}


%
\IEEEpeerreviewmaketitle

\section{Introduction}
\label{intro}

\IEEEPARstart{I}{n} recent years, deep learning has made significant breakthroughs in remote sensing and Earth observation \cite{zhu2017deep,weng2024will}, with widespread application in image interpretation tasks such as scene classification \cite{ma2021scenenet}, object detection \cite{xia2018dota}, and semantic segmentation \cite{wang2022unetformer}. While these achievements are certainly attributed to the increasingly powerful learning capabilities of deep neural networks \cite{zhang2022artificial}, the emergence of large-scale, high-quality remote sensing image datasets is also indispensable \cite{schmitt2023there}. However, such reliance on large-scale datasets inevitably introduces considerable storage and computational overheads \cite{sun2021research}. Therefore, reducing the dependence on large-scale real datasets while maintaining model performance has become a key issue that needs to be addressed for remote sensing image interpretation.

\begin{figure}[t]
  \centering
  \includegraphics[width=\linewidth]{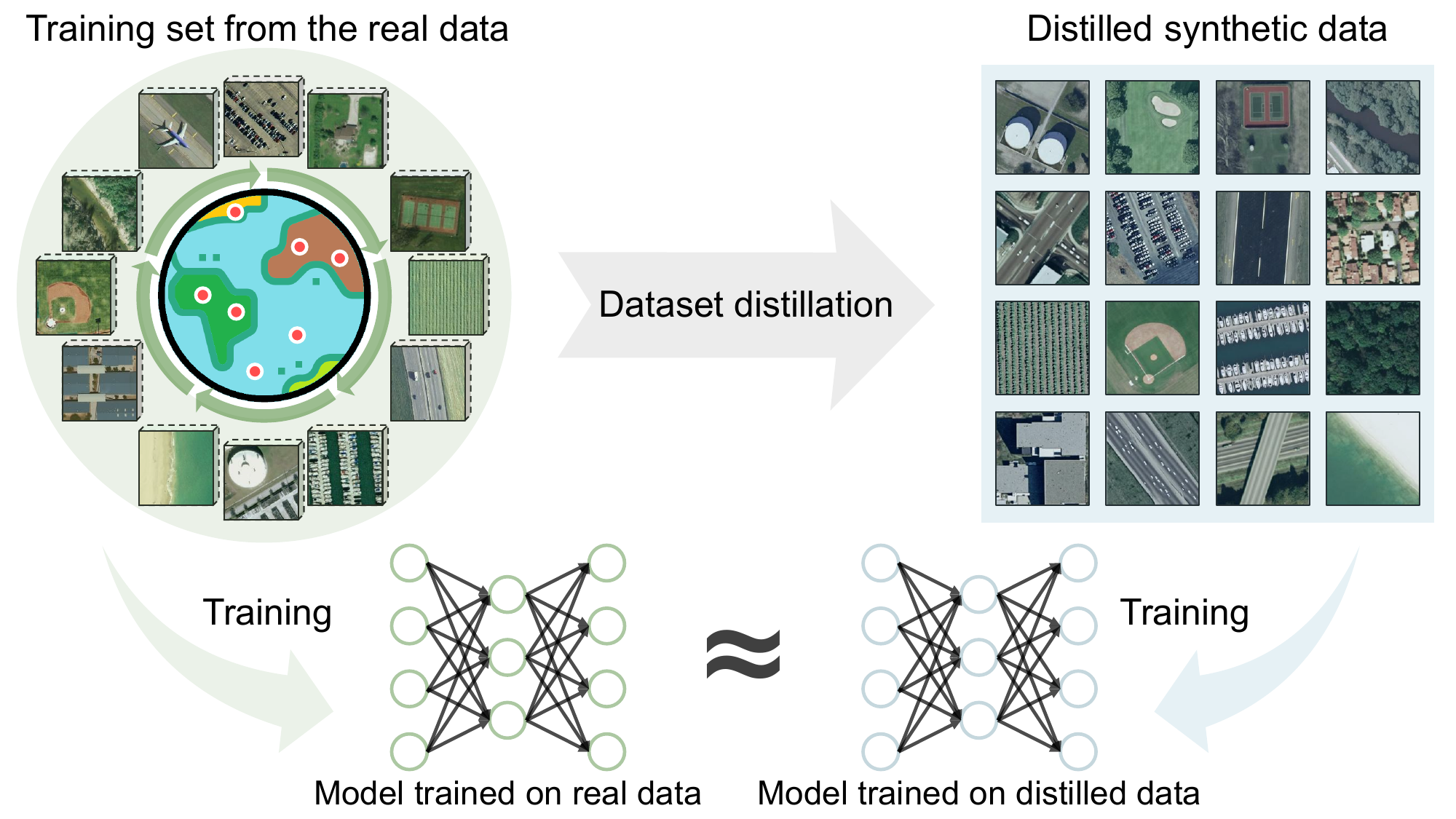}
  \caption{Overview of the motivation for this work. Through dataset distillation, large-scale remote sensing datasets can be condensed into compact synthetic datasets that enable models to achieve performance comparable to training on the original full dataset.}
\label{fig:intro}
\end{figure}

To alleviate this challenge, dataset distillation, an emerging research direction, has recently gained increasing attention \cite{lei2023comprehensive,yu2023dataset}. As shown in Figure~\ref{fig:intro}, the core idea of dataset distillation is to generate a small number of representative synthetic samples to replace the original large-scale training set to train the deep learning models, thereby reducing data storage and computing costs while maintaining model performance \cite{wang2018dataset}. Unlike standard data augmentation, which aims to expand the training set to improve model robustness, dataset distillation pursues the opposite goal, i.e., to identify the ``distilled essence'' or a minimal set of synthetic samples required for a model to learn effectively. Although previous studies have demonstrated the effectiveness of dataset distillation in the field of natural image classification \cite{gu2024efficient,zhao2025taming}, its direct application to remote sensing image interpretation tasks is still challenging. Compared with natural images, remote sensing images not only cover rich and complex object categories, but also exhibit diverse imaging conditions, visual styles, and semantic contexts \cite{zhang2016deep}, which makes it difficult to generate distilled samples that are both discriminative and diverse. Therefore, to achieve more efficient remote sensing dataset distillation, the key is to fully exploit the rich semantic information contained in remote sensing data to accurately guide the synthesis process.

One natural idea is to use generative models to synthesize realistic remote sensing images and construct a condensed virtual training set \cite{xu2023txt2img}. Some studies have attempted to design generative adversarial networks (GANs) \cite{zhao2021text}, auto-regressive models \cite{xu2023txt2img}, or diffusion models \cite{tang2024crs} for remote sensing text-to-image generation. However, these methods often focus on fitting the overall visual style and texture structure of remote sensing images, making it difficult to ensure the diversity of generated data. Furthermore, existing methods usually rely on metrics such as the Inception Score (IS) \cite{salimans2016improved} and Fr\'{e}chet Inception Distance (FID) \cite{heusel2017gans} to measure image realism, but lack specialized optimization designs for downstream interpretation tasks. As a result, even if the generated images perform well in terms of visual quality metrics, the corresponding virtual training set may not effectively improve the performance of the trained classifier \cite{xu2023txt2img}.

Building on these observations, we propose a discriminative prototype-guided diffusion (DPD) model for remote sensing dataset distillation. Unlike existing image generation approaches that primarily focus on visual realism, DPD aims to synthesize compact samples with both semantic fidelity and discriminative quality. Specifically, we first pretrain a latent diffusion model on the original remote sensing dataset and cluster the training samples in the latent space to identify representative and diverse prototypes for each category. We then construct hyperspherical semantic anchors around these prototypes to guide the reverse denoising trajectory. Finally, multiple candidates are generated for each prototype and ranked by a latent classifier using a logit-margin criterion to select the most discriminative candidates for dataset distillation. Despite its simplicity, the proposed method can produce realistic, diverse, and discriminative samples for downstream model training. The main contributions of this paper are summarized as follows:

\begin{enumerate}
    \item We systematically analyze the feasibility of dataset distillation for remote sensing data for the first time. Our results demonstrate the potential of training interpretation models with fully synthetic samples.    
    \item We propose a novel discriminative prototype-guided diffusion (DPD) model for remote sensing dataset distillation, which exploits representative latent prototypes as semantic anchors to guide the reverse denoising trajectory for synthesizing compact and diverse samples.    
    \item We further design a discriminative latent candidate selection strategy that ranks multiple generated candidates with a latent classifier and selects candidates with higher logit margins, thereby improving the discriminative quality of the distilled dataset.
    \item Extensive experiments on three remote sensing scene classification datasets demonstrate that the proposed DPD can distill high-quality, diverse, and discriminative samples for downstream model training compared to state-of-the-art methods.
\end{enumerate}

The rest of this paper is organized as follows. Section~\ref{related_work} reviews related work, Section~\ref{method} details the proposed DPD model, Section~\ref{exp} presents the datasets and experimental results, and Section~\ref{conclusion} concludes the paper with some discussions.

\section{Related Work}
\label{related_work}
This section provides a brief review of data reduction paradigms, along with existing dataset distillation methods and text-to-image generation methods for remote sensing.

\subsection{Data Reduction Paradigms}
Reducing reliance on large-scale training datasets has long been an important problem in efficient machine learning. Existing data reduction strategies mainly include \textit{data pruning}~\cite{sorscher2022beyond}/\textit{coreset selection}~\cite{bachem2015coresets}, which retain a compact subset of real training samples, and \textit{dataset distillation}, which synthesizes a compact training set to preserve the task-relevant knowledge of the original dataset. Data pruning and coreset selection are intuitive to implement, since the selected samples remain valid real images with preserved labels~\cite{yang2023dataset}. For example, Nogueira et al. introduced six basic coreset selection approaches to identify high-quality subsets for remote sensing semantic segmentation~\cite{nogueira2026core}. Wei et al. proposed a training-free data pruning method for remote sensing diffusion foundation models, which combines entropy-based filtering and scene-aware clustering to select informative and representative subsets under high pruning ratios~\cite{wei2025rs}. However, the samples in the pruned dataset remain unmodified and are constrained by the original data distribution, which limits the expressive power of the reduced dataset, especially under limited data budgets~\cite{lei2023comprehensive}.

By contrast, instead of retaining existing real samples, dataset distillation aims to construct a compact synthetic training set that preserves the task-relevant knowledge of the original dataset. Therefore, distilled samples are not required to correspond to any individual real image, providing greater flexibility for encoding class-discriminative information and covering intra-class variations under a limited data budget. In practice, data pruning asks which real samples should be retained, whereas dataset distillation asks what synthetic samples should be constructed to best replace the original dataset.

\subsection{Dataset Distillation}
Since its introduction by Wang et al. \cite{wang2018dataset}, existing dataset distillation methods have mainly followed two technical paradigms: \textit{meta-learning-based} methods and \textit{data matching-based} methods \cite{lei2023comprehensive}. In the meta-learning paradigm, the distilled samples are treated as hyperparameters and optimized in a nested bi-level manner, where the inner loop trains a model on the synthetic set and the outer loop updates the synthetic data to minimize the loss on the real dataset \cite{bohdal2020flexible, sucholutsky2021soft}. In contrast, data matching approaches update distilled samples by aligning the influence of synthetic and real data during training, usually in the form of gradients, trajectories, and feature distributions \cite{zhao2021dataset, cazenavette2022dataset, zhao2023dataset}. 

However, the synthetic data generated by these methods are not visually realistic, which has motivated recent studies to incorporate advanced generative models to enhance the realism of distilled data \cite{zhao2025taming}. For example, Wang et al. utilized a generative model to store the knowledge of the target dataset, where distillation is achieved by minimizing the logit discrepancies between real and synthetic images \cite{wang2024dim}. Su et al. proposed a dataset distillation framework based on a disentangled diffusion model, which enforces consistency between real and synthetic image spaces to improve cross-architecture generalization capabilities \cite{su2024d}. Zhao et al. further proposed a diffusion-based framework that maps real data into a high-normality Gaussian domain to preserve structural information and generate representative latents for distilled datasets \cite{zhao2025taming}.

While dataset distillation has shown effectiveness in natural image classification, its direct application to remote sensing image interpretation remains challenging due to complex semantic and spatial patterns \cite{zhang2016deep}. 

\subsection{Text-to-Image Generation for Remote Sensing}
Driven by recent advances in generative models, text-to-image generation techniques have achieved great success in the field of natural images \cite{ramesh2021zero} and are being increasingly explored for remote sensing applications \cite{li2024vision}. By conditioning on natural language descriptions, such models can synthesize remote sensing images with specific semantic content, providing a new strategy for data augmentation in interpretation tasks \cite{xu2023txt2img}.

Early work in this field focused on using GANs to synthesize remote sensing imagery. One representative example is StrucGAN, a multistage GAN that incorporates structural information from an unsupervised segmentation module to guide the generator, thereby producing structurally consistent images from text descriptions \cite{zhao2021text}. Given the limitations of GANs in terms of training stability and generative diversity, some research has explored auto-regressive models. For example, Xu et al. proposed Txt2Img-MHN, which adopts modern Hopfield networks to learn hierarchical vision prototypes from text-image embeddings in an auto-regressive manner \cite{xu2023txt2img}. In recent years, diffusion models have also been explored for remote sensing data generation due to their breakthrough performance in improving image quality and semantic consistency \cite{liu2024diffusion}. Some representative methods include Crs-diff \cite{tang2024crs}, DiffusionSat \cite{khannadiffusionsat}, GeoRSSD \cite{zhang2024rs5m}, and  Text2Earth \cite{liu2025text2earth}.

So far, most existing work on remote sensing text-to-image generation has primarily focused on visual realism, with results usually evaluated by visual quality metrics such as IS and FID. However, since the discriminative performance of generated samples is not explicitly optimized during training, high visual quality does not always translate into improved classification performance in downstream applications \cite{xu2023txt2img}. Therefore, balancing the fidelity and discriminative quality of synthesized samples still remains a challenge.

\section{Methodology}
\label{method}
The core idea of the proposed discriminative prototype-guided diffusion (DPD) is to use category prototypes as latent-space semantic anchors for remote sensing dataset distillation. Specifically, a latent diffusion model is first pretrained on the original remote sensing dataset to learn the class-conditional data distribution. Then, representative prototypes are extracted for each category, and hyperspherical semantic anchors are constructed to guide the reverse diffusion trajectory. Finally, multiple candidates are generated for each prototype and ranked by a latent classifier using a logit-margin criterion, with the most discriminative candidate for each prototype decoded as the final distilled sample.

\begin{figure*}[t]
  \centering
  \includegraphics[width=\linewidth]{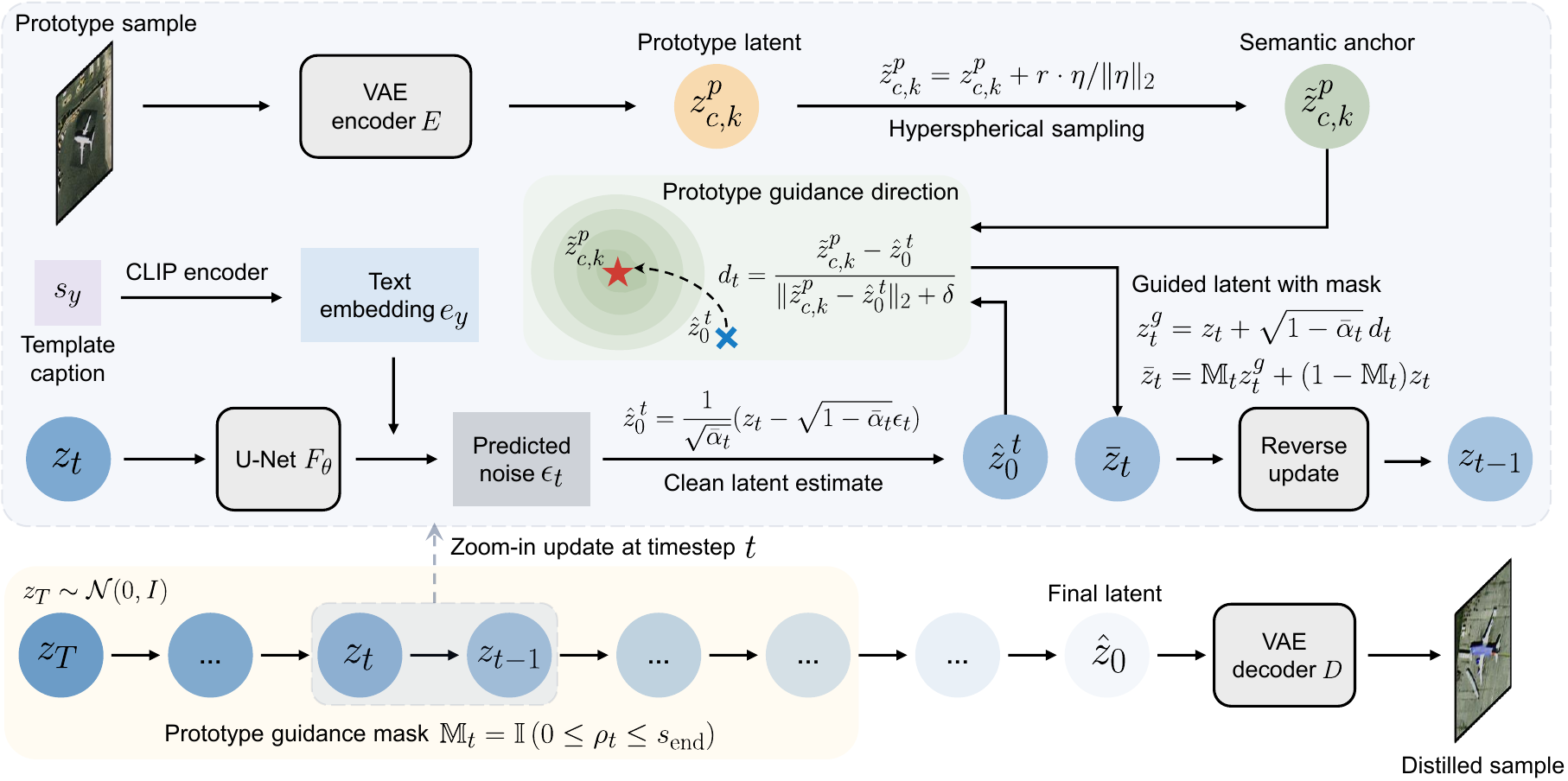}
  \caption{Overview of the proposed prototype-guided latent denoising process. Given a prototype sample, DPD constructs a semantic anchor via hyperspherical sampling in the latent space. During reverse diffusion, the clean latent estimate at each guided timestep is adjusted toward the semantic anchor through a prototype guidance direction, and the resulting window-masked latent is used for the reverse update. The final denoised latent is decoded by the VAE decoder to obtain the distilled sample.}
\label{fig:dpd}
\end{figure*}

\subsection{Problem Formulation}
The objective of dataset distillation is to synthesize a compact distilled dataset $\mathcal{D} = \{(x_i, y_i)\}_{i=1}^{N_\mathcal{D}}$ from a large-scale real dataset $\mathcal{R} = \{(x_i, y_i)\}_{i=1}^{N_\mathcal{R}}$, where $x_i$ denotes an image sample, $y_i$ is its corresponding class label, and typically $N_\mathcal{D} \ll N_\mathcal{R}$. The distilled dataset $\mathcal{D}$ is expected to preserve the essential knowledge of $\mathcal{R}$, such that models trained on $\mathcal{D}$ can generalize to the unseen test dataset $\mathcal{T}$ in the real domain with performance comparable to those trained on the original large-scale dataset $\mathcal{R}$.

\subsection{Latent Diffusion Pretraining}
Diffusion models are a class of probabilistic generative models that progressively transform data into noise and learn to invert this process to recover the original data distribution \cite{ho2020denoising}. Formally, the forward diffusion defines a Markov chain of length $T$ that gradually corrupts a real image sample $x$ with Gaussian noise to obtain intermediate states:
\begin{equation}
\label{diffusion_pixel}
x_t = \sqrt{\bar{\alpha}_t}x + \sqrt{1-\bar{\alpha}_t}\epsilon,
\end{equation}
where $\epsilon \sim \mathcal{N}(0, I)$ denotes a standard Gaussian noise. The term $\bar{\alpha}_t = \prod_{s=1}^{t}\alpha_s$ controls the noise level at timestep $t \in [1, T]$, where $\alpha_s = 1 - \beta_s$, and $\{\beta_s\}_{s=1}^T$ is a series of noise scales predefined by the diffusion process that determine the amount of Gaussian noise added at each step. Intuitively, $\sqrt{\bar{\alpha}_t}$ represents the proportion of the original image that remains after $t$ diffusion steps, while $\sqrt{1-\bar{\alpha}_t}$ quantifies the accumulated noise component. As $t$ increases, $\bar{\alpha}_t$ gradually decreases towards $0$, and at the final step $t=T$, the sample $x_T$ becomes approximately pure Gaussian noise.

Inspired by the work in \cite{rombach2022high}, our framework performs diffusion in a compact latent space learned by a variational autoencoder (VAE) to improve computational efficiency. Specifically, the VAE encoder $E$ first maps a real image $x$ into a latent representation $z_0 = E(x)$. The forward diffusion in Eq. \eqref{diffusion_pixel} is then reformulated as
\begin{equation}
\label{diffusion_latent}
z_t = \sqrt{\bar{\alpha}_t}z_0 + \sqrt{1-\bar{\alpha}_t}\epsilon.
\end{equation}

During the reverse diffusion, a U-Net-based denoising network $F_\theta$ is trained to predict the noise component from the noisy latent $z_t$ at each timestep $t$: 
\begin{equation}
\label{noise_pred}
\epsilon_t = F_\theta(z_t, t, e_y),
\end{equation}
where $e_y = E_\text{txt}(s_y)$ denotes the text embedding obtained by applying a CLIP text encoder $E_\text{txt}$ to the template caption $s_y$, which provides semantic guidance during the denoising process \cite{radford2021learning}. Given the image-label pairs in the original dataset $\mathcal{R}$, the template caption $s_y$ is defined as ``A satellite image of \texttt{<class name>}'', where \texttt{<class name>} denotes the category name associated with label $y$. 

The diffusion model is trained by minimizing the mean squared error (MSE) between the predicted and true noise:
\begin{equation}
\label{diffusion_loss}
\mathcal{L}_{\mathrm{diff}} = 
\mathbb{E}_{t \sim \text{Uniform}(1,T),\,z_0,\,\epsilon,\,y}
\big[\|\epsilon - F_\theta(z_t, t, e_y)\|^2\big].
\end{equation}

\subsection{Latent Prototype Extraction}
\label{prototype}

Due to variations in imaging illumination, viewpoint, scale, and spatial layout, remote sensing images from the same category often exhibit significant intra-class diversity. To preserve such intra-class diversity and maintain the semantic consistency of each category under a limited distillation budget, we extract multiple representative prototypes for each category in the latent space.

Formally, given the original dataset  $\mathcal{R} = \{(x_i, y_i)\}_{i=1}^{N_\mathcal{R}}$, each image $x_i$ is encoded by the VAE encoder into a latent representation $z_i = E(x_i)$. For each class $c \in \{1, \dots, C\}$, we collect its latent features as $\mathcal{Z}_c = \{z_i \mid y_i = c\}$. To model the intra-class variability, we perform K-means clustering on $\mathcal{Z}_c$ and partition it into $K$ sub-clusters $\mathcal{Z}_{c,k} = \{z_i \mid \text{cluster}(z_i) = k\}$, where $k \in \{1, \dots, K\}$.
The centroid of each cluster is computed as $
\mu_{c,k} = \sum_{z_i \in \mathcal{Z}_{c,k}} z_i/|\mathcal{Z}_{c,k}|$.

Within each cluster $\mathcal{Z}_{c,k}$, we select a representative prototype sample that balances intra-cluster compactness and inter-cluster separability. Specifically, for each latent feature $z_i \in \mathcal{Z}_{c,k}$, we define a prototype margin metric as
\begin{equation}
\label{prototype_margin}
M_{\mathrm{proto}}(z_i) =
\min_{j \neq k}\|z_i - \mu_{c,j}\|_2
-
\|z_i - \mu_{c,k}\|_2,
\end{equation}
where the first term measures the minimum distance from $z_i$ to the centroids of other clusters within the same class, and the second term measures its distance to the centroid of its assigned cluster. A larger value indicates that $z_i$ is compact with respect to its own cluster while being well separated from other intra-class clusters.

The prototype for the $k$-th cluster is then selected as the sample that maximizes this margin:
\begin{equation}
\label{margin-selection}
z_{c,k}^{p} = \arg\max_{z_i \in \mathcal{Z}_{c,k}} M_{\mathrm{proto}}(z_i).
\end{equation}
The criterion in Eq.~\eqref{margin-selection} encourages each prototype to be representative of its local cluster while remaining sufficiently distinct from other clusters. The resulting prototype set for class $c$ is denoted as $\mathcal{P}_c = \{z_{c,k}^{p}\}_{k=1}^{K}$.

\subsection{Prototype-guided Latent Denoising}
\label{sec:prototype_guided_denoising}
Without proper guidance, diffusion models often oversample the most representative but redundant samples, which limits the diversity in the distilled dataset \cite{gu2024efficient}. To address this issue, we exploit latent prototypes as semantic anchors to guide the denoising trajectory, as shown in Figure~\ref{fig:dpd}.  

Specifically, given a prototype latent $z_{c,k}^{p}$, we sample a semantic anchor on a hypersphere centered at the prototype $\tilde{z}_{c,k}^{p} = z_{c,k}^{p} + r \cdot\eta/\|\eta\|_2$, where $\eta\sim \mathcal{N}(0,I)$ denotes a Gaussian random vector, $d_{\mathrm{emb}}$ denotes the latent embedding dimension, and $r=1/\sqrt{d_{\mathrm{emb}}}$ controls the radius of the hypersphere. The reverse diffusion process starts from a Gaussian noise $z_T \sim \mathcal{N}(0,I)$, and proceeds with the pretrained diffusion model under the text embedding condition $e_y$. At timestep $t$, the denoising network predicts the noise component $\epsilon_t = F_\theta(z_t,t,e_y)$, from which the current clean latent estimate is obtained by the Denoising Diffusion Probabilistic Models (DDPM) parameterization \cite{ho2020denoising}:
\begin{equation}
\label{eq:predicted_clean_latent}
\hat{z}_0^{\,t}
=
\frac{1
}
{\sqrt{\bar{\alpha}_t}}
(z_t-\sqrt{1-\bar{\alpha}_t}\epsilon_t).
\end{equation}

We then compute the normalized direction from the current clean estimate to the semantic anchor $\tilde{z}_{c,k}^{p}$:
\begin{equation}
\label{eq:guidance_direction}
d_t
=
\frac{
\tilde{z}_{c,k}^{p}-\hat{z}_0^{\,t}
}
{
\|\tilde{z}_{c,k}^{p}-\hat{z}_0^{\,t}\|_2+\delta
},
\end{equation}
where $\delta$ is a small constant for numerical stability. 

Accordingly, the prototype-guided latent can be defined as
\begin{equation}
\label{eq:guided_latent}
z_t^{g}
=
z_t
+
\sqrt{1-\bar{\alpha}_t}\,d_t,
\end{equation}
where $\sqrt{1-\bar{\alpha}_t}$ adaptively scales the guidance according to the noise level.

To prevent excessive constraint throughout the entire denoising trajectory, prototype guidance is activated only within a predefined guidance window. Let $\rho_t=(T-t)/T$ denote the normalized reverse progress. The guidance mask at timestep $t$ is defined as
\begin{equation}
\label{eq:guidance_window}
\mathbb{M}_t
=
\mathbb{I}
\left(
0
\leq
\rho_t
\leq
s_{\mathrm{end}}
\right),
\end{equation}
where $\mathbb{I}(\cdot)$ denotes the indicator function, and $s_{\mathrm{end}}$ is the end ratio of the guidance window (see Section~\ref{sec:ablation} for a detailed analysis of hyperparameter $s_{\mathrm{end}}$). The latent used for the reverse update is then given by
\begin{equation}
\label{eq:actual_guided_latent}
\bar{z}_t
=
\mathbb{M}_t z_t^{g}
+
(1-\mathbb{M}_t)z_t.
\end{equation}

At each timestep, $\bar{z}_t$ is fed into the standard reverse diffusion scheduler to obtain the next denoised latent $z_{t-1}$. Repeating this update from $t=T$ to $t=1$ yields the final denoised latent $\hat{z}_0$, which is denoted as $z_{c,k}^{\mathrm{gen}}$ for the corresponding prototype $z_{c,k}^{p}$. The VAE decoder then reconstructs the distilled sample as $\hat{x}_{c,k}=D(z_{c,k}^{\mathrm{gen}})$.

\subsection{Discriminative Latent Candidate Selection}
\label{sec:discriminative_selection}
While prototype-guided denoising produces semantically aligned samples with more diversity, their discriminative quality may still vary due to stochastic initialization. To identify samples with higher discriminative quality, we further train a latent classifier $f_\phi: z \mapsto \mathbb{R}^{C}$ on the latent representations of the original dataset $\mathcal{R}$  with a standard cross-entropy loss:
\begin{equation}
\label{eq:latent_classifier_loss}
\mathcal{L}_{\mathrm{cls}}
=
\mathbb{E}_{(x_i,y_i)\sim \mathcal{R}}
\left(
-\log
\frac{
\exp(f_\phi(E(x_i))_{y_i})
}
{
\sum_{j=1}^{C}\exp(f_\phi(E(x_i))_j)
}
\right).
\end{equation}

For each prototype $z_{c,k}^{p}$, we repeat prototype-centered hypersphere sampling and prototype-guided denoising $B$ times (see Section~\ref{sec:ablation} for a detailed analysis of hyperparameter $B$), yielding a candidate set $\mathcal{C}_{c,k}=\{z_{c,k}^{(b)}\}_{b=1}^{B}$. 
The discriminative quality of each candidate is evaluated by a logit-margin score using the trained latent classifier $f_\phi$:
\begin{equation}
\label{eq:classification_margin}
S(z)
=
f_\phi(z)_c
-
\max_{j\neq c} f_\phi(z)_j,
\end{equation}
where a larger margin value indicates stronger confidence for the target class $c$ over competing classes. We then select the candidate for each prototype with the largest margin value $z_{c,k}^{*}=\arg\max_{z\in \mathcal{C}_{c,k}}S(z)$, and decode it into the distilled image $x_{c,k}^{*}=D(z_{c,k}^{*})$. The final distilled dataset is constructed by collecting the selected sample from each prototype:
\begin{equation}
\label{eq:distilled_dataset}
\mathcal{D}
=
\left\{
(x_{c,k}^{*},c)
\mid
c\in\{1,\dots,C\},
k\in\{1,\dots,K\}
\right\}.
\end{equation}

The detailed implementation of the dataset distillation process is presented in Algorithm~\ref{alg:dataset_distillation}.

\begin{algorithm}[t]
\caption{Dataset Distillation with DPD}
\label{alg:dataset_distillation}
\KwIn{
Training dataset $\mathcal{R}$; VAE encoder $E$ and decoder $D$;  denoising network $F_\theta$; latent classifier $f_\phi$; prototype number $K$; candidate number $B$; guidance end ratio $s_{\mathrm{end}}$.
}
\KwOut{Distilled dataset $\mathcal{D}$.}

\algcomment{Pretraining} \\
Train $F_\theta$ on $\mathcal{R}$ using Eq.~\eqref{diffusion_loss}. \\
Train $f_\phi$ in latent space on $\mathcal{R}$ using Eq.~\eqref{eq:latent_classifier_loss}. \\
Initialize the distilled dataset $\mathcal{D}\leftarrow\emptyset$. \\

\algcomment{Latent prototype extraction} \\
\For{each class $c\in\{1,\dots,C\}$}{
    Collect latent features  $\mathcal{Z}_c=\{E(x_i)\mid y_i=c\}$. \\
    Cluster $\mathcal{Z}_c$ into sub-clusters $\{\mathcal{Z}_{c,k}\}_{k=1}^{K}$ and select prototypes $\{z_{c,k}^{p}\}_{k=1}^{K}$ using Eq.~\eqref{margin-selection}. \\

    \For{each prototype $z_{c,k}^{p}$}{
        Initialize the candidate set $\mathcal{C}_{c,k}\leftarrow\emptyset$. \\

        \algcomment{Prototype-guided denoising} \\
        \For{$b=1$ to $B$}{
            Sample a hyperspherical semantic anchor $\tilde{z}_{c,k}^{p}=z_{c,k}^{p}+r\cdot\eta/\|\eta\|_2$, $\eta\sim\mathcal{N}(0,I)$. \\
            Generate candidate latent $z_{c,k}^{(b)}$ with Eqs.~\eqref{eq:predicted_clean_latent}--\eqref{eq:actual_guided_latent} and add $z_{c,k}^{(b)}$ to $\mathcal{C}_{c,k}$. \\
        }

        \algcomment{Candidate selection} \\
        Select $z_{c,k}^{*}=\arg\max_{z\in\mathcal{C}_{c,k}}S(z)$ with Eq.~\eqref{eq:classification_margin}. \\
        Add $(D(z_{c,k}^{*}),c)$ to $\mathcal{D}$. \\
    }
}

\Return{$\mathcal{D}$}
\end{algorithm}

\section{Experiments}
\label{exp}
\subsection{Data Descriptions}
We conduct experiments on three benchmark remote sensing scene classification datasets, including UC Merced (UCM) \cite{yang2010bag}, aerial image dataset (AID) \cite{aid}, and NWPU-RESISC45 \cite{cheng2017remote}.

\textbf{UCM} contains 2,100 aerial scene images across 21 land-use categories, with 100 images per class. All images are RGB with a resolution of $256\times256$ pixels (0.3 m spatial resolution) and are derived from USGS National Map aerial orthoimagery. 

\textbf{AID} is a large-scale dataset collected from Google Earth. It is made up of 10,000 RGB images across 30 aerial scene categories. All images are annotated by remote sensing experts and provided at a fixed resolution of 600$\times$600 pixels, with spatial resolutions ranging from 0.5 m to 8 m. The number of images per class varies between 220 and 420.

\textbf{NWPU-RESISC45} contains 31,500 remote sensing images across 45 scene categories, with 700 RGB images per class. It was collected by remote sensing experts from Google Earth, covering over 100 countries and regions. All images are provided at a resolution of $256\times256$ pixels, with spatial resolutions ranging from 0.2 m to 30 m. 

For UCM and AID, we randomly split each dataset into 50\% for the training set $\mathcal{R}$ and the remaining 50\% for the testing set $\mathcal{T}$. For NWPU-RESISC45, we follow the official protocol from the original paper, where 27,000 images are used as the training set $\mathcal{R}$ and the remaining 4,500 images serve as the testing set $\mathcal{T}$.

\subsection{Experimental Settings}
For each dataset distillation method evaluated in this study, we consider multiple images-per-class (IPC) settings, i.e., $\text{IPC}\in\{5, 10, 15, 20\}$. Under each setting, the distillation model is trained on the real training set $\mathcal{R}$ to synthesize IPC synthetic samples per class, which together form the distilled dataset $\mathcal{D}$. 

To assess the quality of the distilled samples produced by each method, we train a classification network using the distilled dataset $\mathcal{D}$ and report the overall accuracy (OA) on the real test set $\mathcal{T}$. The evaluated network architectures include VGG16~\cite{vgg}, Inception-v3~\cite{inception}, ResNet18~\cite{he2016deep}, and DenseNet121~\cite{densenet}. Each experiment is repeated five times with random seeds, and we report the mean accuracy along with the standard deviation.

\subsection{Implementation Details}
\textbf{Latent diffusion pretraining.} We adopt Stable Diffusion 2 \cite{rombach2022high} as the text-to-image generative backbone for dataset distillation, which provides pretrained weights for VAE encoder $E$ and decoder $D$. The CLIP text encoder $E_{\text{txt}}$ is instantiated with OpenCLIP ViT-H-14 \cite{ilharco_gabriel_2021_5143773} pretrained on LAION-2B \cite{schuhmann2022laionb}. Both the VAE and CLIP text encoder are kept frozen throughout training. To enable more efficient adaptation, we fine-tune the U-Net denoising network $F_\theta$ with LoRA (Low-Rank Adaptation) \cite{hu2022lora}, initialized from Text2Earth \cite{liu2025text2earth}. The model is optimized using AdamW \cite{loshchilov2017decoupled} with a learning rate of $5e-4$, batch size of 8, and 20,000 training steps at the resolution of $256\times256$.

\begin{figure}[t]
  \centering  
\includegraphics[width=\linewidth]{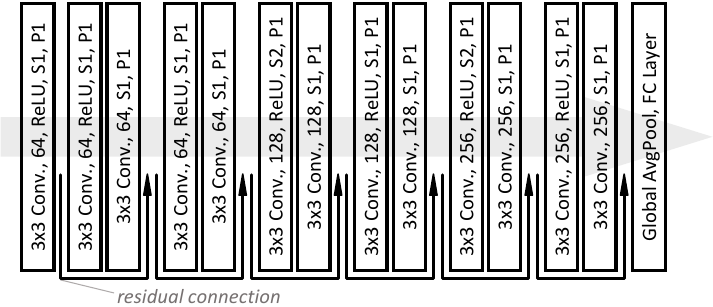}
  \caption{Detailed architecture of the latent classifier used in this study. The network takes VAE latent features as input and consists of six residual blocks, followed by global average pooling and a fully connected layer for scene classification. ``S'' and ``P'' denote stride and padding, respectively.
}
\label{fig:latent_classifier}
\end{figure}

\textbf{Latent classifier pretraining}. To evaluate the discriminative quality of generated latent candidates, we pretrain a lightweight latent classifier $f_\phi$ on VAE-encoded remote sensing images for each dataset. Specifically, each image is resized to $256\times256$ and encoded by the frozen VAE encoder $E$ into latent posterior statistics. The classifier takes the resulting 4-channel latent features as input and performs scene classification. Its detailed architecture is illustrated in Fig.~\ref{fig:latent_classifier}. The classifier is optimized using AdamW with a learning rate of $1e-3$, batch size of 128, and 200 training epochs.

\textbf{Prototype-guided latent denoising}. For each category in $\mathcal{R}$, we perform K-means clustering in the VAE latent space, with the number of clusters set to $K=\mathrm{IPC}$ under the dataset distillation setting. In semantic anchor sampling, $d_{\mathrm{emb}}=4096$ denotes the latent embedding dimension. The end ratio of the guidance window $s_{\mathrm{end}}$ in Eq. \eqref{eq:guidance_window} is set to 0.8, and the number of candidates $B$ for discriminative latent candidate selection is set to 5. Detailed analyses of these two hyperparameters are provided in Section~\ref{sec:ablation}.

All experiments in this study are conducted on a cluster with a single NVIDIA A100 GPU.

\begin{table*}[t]
\caption{Dataset distillation performance in terms of OA (\%) on the UCM dataset across different classification networks.}
\label{tab:ucm}
\centering
\resizebox{\linewidth}{!}{%
\begin{tabular}{cc|cccccc>{\columncolor{gray!7}}c|c}
\toprule
Model & IPC (Ratio) & MinimaxDiff~\cite{gu2024efficient} & Text2Earth~\cite{liu2025text2earth} & CaO$_2$~\cite{wang2025cao2} & DiT-IGD~\cite{chen2025influence} & ImS$^3$~\cite{wang2026ims3} & ManifoldGD~\cite{roy2026manifoldgd} & DPD (ours) & Full (100\%) \\
\midrule
\multirow{4}{*}{VGG16}
& 5 (10\%)  & 56.06$\pm$2.93 & 39.45$\pm$2.51 & 61.20$\pm$1.13 & 64.55$\pm$3.35 & 64.63$\pm$2.84 & \textbf{73.58$\pm$1.40} & 69.70$\pm$2.71 & \multirow{4}{*}{94.29$\pm$0.34} \\
& 10 (20\%) & 67.96$\pm$2.44 & 44.55$\pm$4.04 & 70.63$\pm$0.95 & 68.74$\pm$3.05 & 71.64$\pm$3.61 & 77.43$\pm$1.68 & \textbf{79.24$\pm$2.21} & \\
& 15 (30\%) & 69.90$\pm$1.77 & 42.19$\pm$3.33 & 73.47$\pm$1.45 & 66.91$\pm$2.24 & 73.66$\pm$4.14 & 81.14$\pm$1.35 & \textbf{81.73$\pm$1.01} & \\
& 20 (40\%) & 71.81$\pm$1.77 & 42.67$\pm$2.62 & 74.76$\pm$1.79 & 65.16$\pm$2.17 & 74.23$\pm$1.66 & \textbf{82.25$\pm$0.40} & 81.54$\pm$3.63 & \\
\midrule
\multirow{4}{*}{Inception-v3}
& 5 (10\%)  & 56.19$\pm$3.06 & 65.12$\pm$1.75 & 59.24$\pm$4.24 & 80.44$\pm$1.91 & 79.18$\pm$0.83 & 82.21$\pm$1.55 & \textbf{84.57$\pm$1.43} & \multirow{4}{*}{95.62$\pm$0.31} \\
& 10 (20\%) & 66.80$\pm$2.62 & 68.84$\pm$2.17 & 72.00$\pm$0.98 & 83.47$\pm$1.82 & 82.76$\pm$1.48 & 85.64$\pm$1.11 & \textbf{88.36$\pm$0.74} & \\
& 15 (30\%) & 74.63$\pm$1.57 & 68.95$\pm$1.23 & 76.70$\pm$3.30 & 86.27$\pm$1.67 & 87.49$\pm$1.16 & 89.26$\pm$0.94 & \textbf{91.07$\pm$0.78} & \\
& 20 (40\%) & 74.88$\pm$1.74 & 69.92$\pm$1.44 & 79.45$\pm$1.09 & 86.99$\pm$1.25 & 87.73$\pm$1.40 & 89.41$\pm$0.42 & \textbf{91.31$\pm$0.98} & \\
\midrule
\multirow{4}{*}{ResNet18}
& 5 (10\%)  & 55.16$\pm$1.64 & 56.44$\pm$3.39 & 57.87$\pm$1.92 & 73.92$\pm$3.24 & 72.84$\pm$2.57 & 79.52$\pm$0.89 & \textbf{80.10$\pm$0.70} & \multirow{4}{*}{97.90$\pm$0.14} \\
& 10 (20\%) & 64.48$\pm$3.58 & 61.10$\pm$2.09 & 71.66$\pm$0.90 & 81.96$\pm$1.19 & 78.04$\pm$2.55 & 84.70$\pm$1.43 & \textbf{87.41$\pm$1.28} & \\
& 15 (30\%) & 69.31$\pm$4.00 & 58.13$\pm$1.96 & 74.82$\pm$2.58 & 81.73$\pm$1.82 & 83.49$\pm$1.26 & 87.28$\pm$1.73 & \textbf{88.53$\pm$0.56} & \\
& 20 (40\%) & 71.20$\pm$3.31 & 60.30$\pm$2.80 & 76.95$\pm$1.33 & 82.86$\pm$2.01 & 84.44$\pm$0.75 & 88.25$\pm$0.98 & \textbf{90.02$\pm$0.89} & \\
\midrule
\multirow{4}{*}{DenseNet121}
& 5 (10\%)  & 57.64$\pm$3.31 & 56.32$\pm$2.30 & 60.11$\pm$3.86 & 78.19$\pm$2.87 & 77.18$\pm$2.30 & 82.55$\pm$1.12 & \textbf{84.38$\pm$1.71} & \multirow{4}{*}{97.71$\pm$0.32} \\
& 10 (20\%) & 67.31$\pm$1.59 & 61.79$\pm$1.72 & 74.50$\pm$0.96 & 83.68$\pm$1.21 & 82.38$\pm$2.14 & 86.50$\pm$1.50 & \textbf{89.35$\pm$0.98} & \\
& 15 (30\%) & 74.65$\pm$3.10 & 59.20$\pm$2.03 & 78.55$\pm$1.48 & 85.47$\pm$1.75 & 87.39$\pm$0.64 & 90.69$\pm$0.46 & \textbf{91.73$\pm$0.82} & \\
& 20 (40\%) & 76.30$\pm$2.30 & 60.53$\pm$1.88 & 79.94$\pm$0.34 & 87.12$\pm$0.78 & 87.56$\pm$0.91 & 90.99$\pm$0.75 & \textbf{92.40$\pm$0.85} & \\
\bottomrule
\end{tabular}
}
{\scriptsize
\noindent
\begin{minipage}{\linewidth}
\vspace{.3em}
Note: Best results are shown in \textbf{bold}. Here, $\text{Ratio}=\text{IPC} \times C / |\mathcal{R}|$ denotes the proportion of distilled samples to the total number of training samples, where $C$ is the number of classes in the training set $\mathcal{R}$. ``Full'' denotes training on the entire original training set $\mathcal{R}$ and is reported as an upper-bound reference.
\end{minipage}
}
\end{table*}

\begin{table*}[t]
\caption{Dataset distillation performance in terms of OA (\%) on the AID dataset across different classification networks.}
\label{tab:aid}
\centering
\resizebox{\linewidth}{!}{%
\begin{tabular}{cc|cccccc>{\columncolor{gray!7}}c|c}
\toprule
Model & IPC (Ratio) & MinimaxDiff~\cite{gu2024efficient} & Text2Earth~\cite{liu2025text2earth} & CaO$_2$~\cite{wang2025cao2} & DiT-IGD~\cite{chen2025influence} & ImS$^3$~\cite{wang2026ims3} & ManifoldGD~\cite{roy2026manifoldgd} & DPD (ours) & Full (100\%) \\
\midrule
\multirow{4}{*}{VGG16}
& 5 (3\%)   & 33.47$\pm$1.75 & 40.12$\pm$1.95 & 43.12$\pm$2.27 & 57.13$\pm$2.64 & 50.19$\pm$2.71 & 61.13$\pm$2.32 & \textbf{66.62$\pm$0.77} & \multirow{4}{*}{91.37$\pm$0.08} \\
& 10 (6\%)  & 37.18$\pm$1.72 & 39.33$\pm$2.36 & 54.87$\pm$1.47 & 60.68$\pm$1.22 & 53.41$\pm$1.22 & 67.90$\pm$2.76 & \textbf{70.34$\pm$0.76} & \\
& 15 (9\%)  & 38.66$\pm$2.09 & 31.47$\pm$1.73 & 57.42$\pm$1.23 & 60.10$\pm$1.64 & 51.82$\pm$1.93 & 69.47$\pm$2.25 & \textbf{71.12$\pm$1.74} & \\
& 20 (12\%) & 36.98$\pm$1.51 & 27.96$\pm$3.07 & 57.84$\pm$0.88 & 62.38$\pm$1.50 & 51.79$\pm$2.79 & 68.25$\pm$0.97 & \textbf{72.90$\pm$1.10} & \\
\midrule
\multirow{4}{*}{Inception-v3}
& 5 (3\%)   & 31.99$\pm$3.20 & 59.01$\pm$1.00 & 45.96$\pm$4.45 & 68.64$\pm$0.73 & 62.30$\pm$1.74 & 72.17$\pm$1.69 & \textbf{80.26$\pm$0.76} & \multirow{4}{*}{94.62$\pm$0.05} \\
& 10 (6\%)  & 43.50$\pm$4.13 & 61.58$\pm$0.83 & 59.59$\pm$2.81 & 75.11$\pm$1.03 & 67.61$\pm$1.05 & 77.92$\pm$1.11 & \textbf{84.73$\pm$0.59} & \\
& 15 (9\%)  & 46.63$\pm$2.57 & 58.52$\pm$0.87 & 61.81$\pm$3.76 & 75.11$\pm$1.52 & 66.84$\pm$1.38 & 76.42$\pm$2.37 & \textbf{84.36$\pm$1.24} & \\
& 20 (12\%) & 47.94$\pm$1.24 & 55.30$\pm$1.30 & 63.98$\pm$0.95 & 75.82$\pm$1.65 & 68.75$\pm$2.71 & 79.38$\pm$1.32 & \textbf{85.27$\pm$1.08} & \\
\midrule
\multirow{4}{*}{ResNet18}
& 5 (3\%)   & 31.78$\pm$2.51 & 52.56$\pm$1.38 & 44.35$\pm$1.73 & 63.94$\pm$0.89 & 57.56$\pm$2.46 & 66.53$\pm$1.66 & \textbf{75.99$\pm$1.53} & \multirow{4}{*}{95.27$\pm$0.18} \\
& 10 (6\%)  & 39.56$\pm$1.75 & 53.14$\pm$1.04 & 56.79$\pm$1.67 & 70.79$\pm$1.14 & 63.60$\pm$1.14 & 74.78$\pm$1.44 & \textbf{81.58$\pm$0.75} & \\
& 15 (9\%)  & 44.58$\pm$2.71 & 54.25$\pm$1.32 & 61.52$\pm$1.81 & 71.65$\pm$1.82 & 67.20$\pm$2.03 & 76.49$\pm$2.56 & \textbf{83.55$\pm$1.14} & \\
& 20 (12\%) & 45.79$\pm$2.60 & 52.46$\pm$1.11 & 64.62$\pm$1.77 & 74.46$\pm$0.89 & 67.21$\pm$2.27 & 78.18$\pm$1.34 & \textbf{84.56$\pm$0.68} & \\
\midrule
\multirow{4}{*}{DenseNet121}
& 5 (3\%)   & 35.76$\pm$1.94 & 56.57$\pm$0.96 & 49.04$\pm$2.82 & 67.18$\pm$1.62 & 62.28$\pm$2.99 & 70.71$\pm$0.84 & \textbf{80.56$\pm$1.07} & \multirow{4}{*}{94.60$\pm$0.09} \\
& 10 (6\%)  & 45.74$\pm$2.68 & 57.99$\pm$1.01 & 60.83$\pm$0.87 & 73.75$\pm$1.10 & 67.18$\pm$1.53 & 78.76$\pm$1.38 & \textbf{84.92$\pm$0.52} & \\
& 15 (9\%)  & 49.59$\pm$2.70 & 57.74$\pm$0.78 & 64.48$\pm$2.50 & 75.59$\pm$1.28 & 70.69$\pm$1.56 & 80.52$\pm$2.51 & \textbf{85.77$\pm$1.00} & \\
& 20 (12\%) & 50.18$\pm$3.20 & 55.97$\pm$1.20 & 66.67$\pm$1.42 & 77.20$\pm$1.36 & 70.03$\pm$1.36 & 80.95$\pm$0.66 & \textbf{86.22$\pm$0.89} & \\
\bottomrule
\end{tabular}
}
{\scriptsize
\noindent
\begin{minipage}{\linewidth}
\vspace{.3em}
Note: Best results are shown in \textbf{bold}. Here, $\text{Ratio}=\text{IPC} \times C / |\mathcal{R}|$ denotes the proportion of distilled samples to the total number of training samples, where $C$ is the number of classes in the training set $\mathcal{R}$. ``Full'' denotes training on the entire original training set $\mathcal{R}$ and is reported as an upper-bound reference.
\end{minipage}
}
\end{table*}

\begin{table*}[t]
\caption{Dataset distillation performance in terms of OA (\%) on the NWPU dataset across different classification networks.}
\label{tab:nwpu}
\centering
\resizebox{\linewidth}{!}{%
\begin{tabular}{cc|cccccc>{\columncolor{gray!7}}c|c}
\toprule
Model & IPC (Ratio) & MinimaxDiff~\cite{gu2024efficient} & Text2Earth~\cite{liu2025text2earth} & CaO$_2$~\cite{wang2025cao2} & DiT-IGD~\cite{chen2025influence} & ImS$^3$~\cite{wang2026ims3} & ManifoldGD~\cite{roy2026manifoldgd} & DPD (ours) & Full (100\%) \\
\midrule
\multirow{4}{*}{VGG16}
& 5 (0.8\%)  & 20.05$\pm$1.52 & 34.84$\pm$2.64 & 32.34$\pm$1.29 & 41.78$\pm$1.63 & 38.42$\pm$2.16 & 53.98$\pm$0.98 & \textbf{57.01$\pm$1.01} & \multirow{4}{*}{90.52$\pm$0.15} \\
& 10 (1.7\%) & 21.17$\pm$1.13 & 32.39$\pm$1.10 & 35.87$\pm$2.18 & 45.05$\pm$2.15 & 38.88$\pm$1.19 & 57.97$\pm$1.37 & \textbf{61.35$\pm$0.61} & \\
& 15 (2.5\%) & 21.51$\pm$1.32 & 32.56$\pm$0.68 & 37.64$\pm$2.06 & 46.90$\pm$3.00 & 38.62$\pm$0.76 & 58.84$\pm$0.38 & \textbf{62.57$\pm$0.56} & \\
& 20 (3.3\%) & 20.89$\pm$1.06 & 31.79$\pm$0.82 & 39.66$\pm$1.55 & 48.34$\pm$0.64 & 38.35$\pm$1.34 & 60.91$\pm$0.54 & \textbf{62.70$\pm$1.45} & \\
\midrule
\multirow{4}{*}{Inception-v3}
& 5 (0.8\%)  & 23.59$\pm$2.01 & 51.74$\pm$1.94 & 36.80$\pm$2.37 & 55.31$\pm$0.89 & 51.20$\pm$2.03 & 63.48$\pm$0.69 & \textbf{68.98$\pm$1.85} & \multirow{4}{*}{95.15$\pm$0.16} \\
& 10 (1.7\%) & 29.84$\pm$1.45 & 52.45$\pm$1.97 & 41.14$\pm$1.58 & 60.70$\pm$2.50 & 54.86$\pm$0.62 & 68.36$\pm$1.36 & \textbf{72.98$\pm$1.51} & \\
& 15 (2.5\%) & 31.82$\pm$1.17 & 49.49$\pm$2.13 & 44.04$\pm$1.40 & 57.42$\pm$1.32 & 52.51$\pm$0.93 & 68.10$\pm$0.84 & \textbf{74.38$\pm$0.69} & \\
& 20 (3.3\%) & 31.55$\pm$0.95 & 43.27$\pm$1.21 & 45.17$\pm$1.88 & 56.98$\pm$0.98 & 50.87$\pm$2.02 & 66.63$\pm$0.89 & \textbf{74.12$\pm$1.07} & \\
\midrule
\multirow{4}{*}{ResNet18}
& 5 (0.8\%)  & 21.09$\pm$1.62 & 43.50$\pm$1.36 & 35.68$\pm$1.71 & 49.66$\pm$1.24 & 43.90$\pm$2.19 & 59.68$\pm$1.50 & \textbf{66.44$\pm$1.06} & \multirow{4}{*}{95.91$\pm$0.16} \\
& 10 (1.7\%) & 28.70$\pm$1.29 & 45.35$\pm$1.44 & 42.51$\pm$0.92 & 59.81$\pm$1.18 & 52.80$\pm$1.18 & 68.16$\pm$0.61 & \textbf{73.61$\pm$1.08} & \\
& 15 (2.5\%) & 31.84$\pm$2.24 & 44.91$\pm$0.63 & 47.98$\pm$1.93 & 60.45$\pm$1.06 & 55.56$\pm$0.91 & 70.54$\pm$1.39 & \textbf{77.56$\pm$0.64} & \\
& 20 (3.3\%) & 33.25$\pm$2.04 & 43.25$\pm$0.47 & 49.02$\pm$0.83 & 61.23$\pm$1.20 & 55.92$\pm$1.32 & 71.40$\pm$0.77 & \textbf{77.20$\pm$0.28} & \\
\midrule
\multirow{4}{*}{DenseNet121}
& 5 (0.8\%)  & 24.15$\pm$1.32 & 45.50$\pm$1.93 & 36.71$\pm$2.61 & 54.08$\pm$1.27 & 46.29$\pm$1.98 & 63.12$\pm$1.28 & \textbf{70.33$\pm$1.33} & \multirow{4}{*}{95.90$\pm$0.16} \\
& 10 (1.7\%) & 32.89$\pm$2.51 & 47.68$\pm$2.90 & 44.26$\pm$1.38 & 62.79$\pm$1.78 & 56.73$\pm$1.65 & 69.87$\pm$1.05 & \textbf{76.76$\pm$0.63} & \\
& 15 (2.5\%) & 33.96$\pm$1.47 & 46.88$\pm$1.81 & 48.15$\pm$2.19 & 62.39$\pm$0.66 & 57.30$\pm$1.22 & 71.42$\pm$1.08 & \textbf{79.45$\pm$0.65} & \\
& 20 (3.3\%) & 32.80$\pm$0.62 & 45.15$\pm$1.10 & 46.61$\pm$1.07 & 63.10$\pm$1.81 & 58.29$\pm$0.71 & 72.51$\pm$0.96 & \textbf{78.37$\pm$1.15} & \\
\bottomrule
\end{tabular}
}
{\scriptsize
\noindent
\begin{minipage}{\linewidth}
\vspace{.3em}
Note: Best results are shown in \textbf{bold}. Here, $\text{Ratio}=\text{IPC} \times C / |\mathcal{R}|$ denotes the proportion of distilled samples to the total number of training samples, where $C$ is the number of classes in the training set $\mathcal{R}$. ``Full'' denotes training on the entire original training set $\mathcal{R}$ and is reported as an upper-bound reference.
\end{minipage}
}
\end{table*}

\begin{figure*}[t]
\centering
\includegraphics[width=\linewidth]{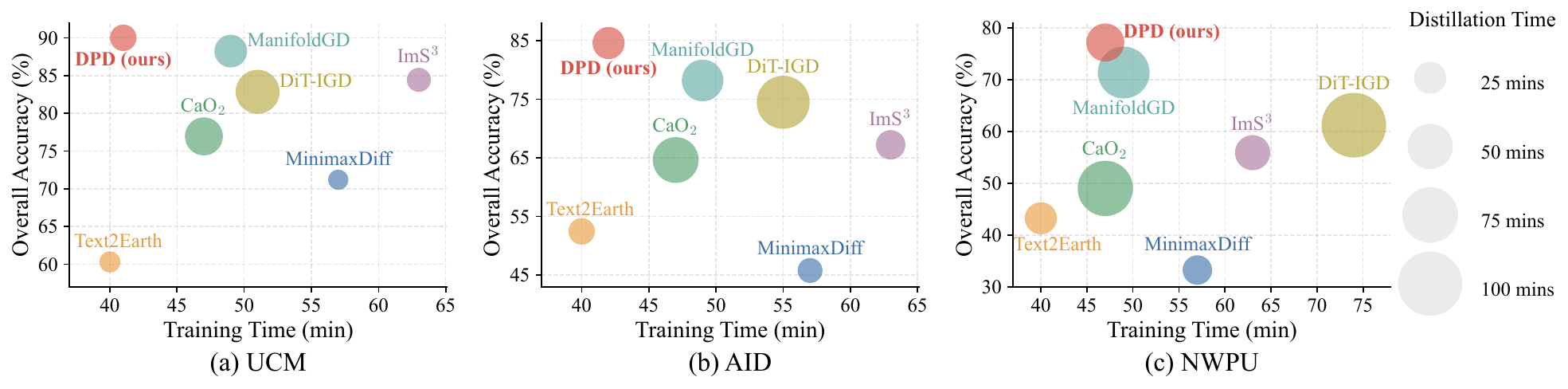}
\caption{Accuracy--efficiency comparison on UCM, AID, and NWPU with IPC $=20$ using ResNet18. The x-axis and y-axis denote the training time of the distillation model and the overall accuracy, respectively. Methods closer to the upper-left corner achieve higher accuracy with shorter distillation-model training time, while the bubble size indicates the time required to generate the distilled dataset.}
\label{fig:time}
\end{figure*}

\subsection{Quantitative Results}
\label{quantitative_results}
In this subsection, we evaluate the quantitative performance of the proposed method along with several recent state-of-the-art approaches. A brief introduction to these methods is given below.

\begin{itemize}
\item MinimaxDiff (CVPR 2024): A generative dataset distillation method that utilizes diffusion models and minimax objectives to synthesize representative and diverse surrogate samples \cite{gu2024efficient}.
\item Text2Earth (GRSM 2025): A large-scale diffusion-based remote sensing text-to-image foundation model with 1.3 billion parameters \cite{liu2025text2earth}.
\item CaO$_2$ (ICCV 2025): A two-stage diffusion-based dataset distillation method that combines probability-informed sample selection with latent refinement \cite{wang2025cao2}.
\item DiT-IGD (ICLR 2025): A diffusion-based dataset distillation method that formulates distillation as controlled diffusion generation via trajectory influence functions \cite{chen2025influence}. 
\item ImS$^3$ (CVPR 2026): A diffusion-based dataset distillation method that combines inversion-guided fine-tuning with selective subgroup sampling \cite{wang2026ims3}.
\item ManifoldGD (CVPR 2026): A diffusion-based dataset distillation method that utilizes hierarchical latent clustering and manifold-consistent guidance \cite{roy2026manifoldgd}.
\end{itemize}

To ensure a fair comparison, all competing methods are fine-tuned using their official implementations under the same training protocol (i.e., the AdamW optimizer with a learning rate of $5e-4$, batch size of 8, and 20,000 training steps at a resolution of $256\times256$).

\textbf{UCM and AID}. Tables~\ref{tab:ucm} and~\ref{tab:aid} report the results on small- to medium-scale remote sensing benchmarks. On UCM, DPD achieves consistently strong performance across different evaluation architectures, obtaining the best results in most settings, with only a few exceptions under VGG16. On the more challenging AID dataset, DPD consistently ranks first across all IPC settings and architectures. Notably, under the low-ratio setting of $\mathrm{IPC}=5$, DPD exceeds the strongest baseline ManifoldGD by more than 8 percentage points on Inception-v3, ResNet18, and DenseNet121. These results indicate that DPD can produce more discriminative and diverse distilled samples even with a highly limited distillation budget.

\textbf{NWPU}. Table~\ref{tab:nwpu} further evaluates scalability on the larger NWPU dataset, where the distilled data account for only 0.8\%--3.3\% of the original training set. Under this more challenging setting, DPD consistently achieves the highest OA across all evaluation networks and IPC settings. In particular, DPD outperforms the strongest baseline ManifoldGD by approximately 8 percentage points on DenseNet121 with $\mathrm{IPC}=15$, demonstrating its clear advantage even when the distilled data account for only 2.5\% of the original dataset. These stable gains across different networks suggest that DPD preserves rich inter-class semantics and intra-class diversity, while also showing strong cross-architecture generalization rather than overfitting to a specific classification model.

\textbf{Accuracy--efficiency comparison}.
Figure~\ref{fig:time} shows the accuracy--efficiency trade-off of different diffusion-based dataset distillation methods on UCM, AID, and NWPU with $\mathrm{IPC}=20$ using ResNet18 as the evaluation network. In each subplot, the horizontal axis denotes the training time of the distillation model (e.g., the time cost for training the diffusion model and the latent classifier in DPD), while the vertical axis denotes OA. The bubble size represents the time required to generate the distilled dataset, i.e., synthesizing $\mathrm{IPC}=20$ samples per class for the target dataset. Therefore, methods closer to the upper-left corner achieve higher accuracy with shorter distillation-model training time, while smaller bubbles indicate lower distilled-data generation cost. As shown in Figure~\ref{fig:time}, DPD consistently lies near the upper-left region across all three datasets, indicating a favorable balance between accuracy and efficiency. Specifically, DPD can be trained within one hour on all datasets, and its distilled-dataset generation time remains below 50 minutes, which is substantially lower than the strong baseline ManifoldGD, especially on the large-scale NWPU dataset. These results demonstrate that DPD improves the discriminative quality of distilled samples without introducing excessive computational overhead.

\subsection{Quality Evaluation of Distilled Samples}

\begin{table*}[t]
\caption{Quality evaluation of distilled samples using CLIP Score, CLIP zero-shot classification OA (\%), FID Score, NN-SSIM, and IC-Coverage.}
\label{tab:metric}
\centering
\resizebox{\linewidth}{!}{%
\begin{tabular}{cc|cccccc>{\columncolor{gray!7}}c}
\toprule
Dataset & Metric 
& MinimaxDiff~\cite{gu2024efficient}
& Text2Earth~\cite{liu2025text2earth}
& CaO$_2$~\cite{wang2025cao2}
& DiT-IGD~\cite{chen2025influence}
& ImS$^3$~\cite{wang2026ims3}
& ManifoldGD~\cite{roy2026manifoldgd}
& DPD (ours) \\
\midrule
\multirow{5}{*}{UCM} 
& CLIP Score ($\times$100) $\uparrow$ & 24.11$\pm$0.11 & 25.50$\pm$0.05 & 24.24$\pm$0.09 & 25.78$\pm$0.04 & 25.69$\pm$0.05 & 25.83$\pm$0.06 & \textbf{26.27$\pm$0.06} \\
& CLIP Zero-shot OA $\uparrow$ & 49.09$\pm$1.24 & 77.19$\pm$1.57 & 52.29$\pm$1.69 & 89.86$\pm$1.41 & 84.33$\pm$1.04 & 89.05$\pm$0.72 & \textbf{90.71$\pm$0.91} \\
& FID Score $\downarrow$ & 208.87$\pm$3.51 & 200.47$\pm$3.70 & 182.11$\pm$6.00 & 75.97$\pm$1.88 & 82.39$\pm$0.68 & \textbf{67.89$\pm$1.72} & 74.58$\pm$0.99 \\
& NN-SSIM ($\times$100) $\downarrow$ & \textbf{18.37$\pm$0.05} & 29.94$\pm$0.17 & 27.62$\pm$0.34 & 31.65$\pm$0.29 & 24.27$\pm$0.23 & 29.12$\pm$0.19 & 31.09$\pm$0.11 \\
& IC-Coverage ($\times$100) $\uparrow$ & 73.38$\pm$0.21 & 72.17$\pm$0.53 & 76.74$\pm$0.06 & 78.46$\pm$0.11 & 78.11$\pm$0.11 & \textbf{79.66$\pm$0.06} & 79.38$\pm$0.22 \\
\midrule
\multirow{5}{*}{AID} 
& CLIP Score ($\times$100) $\uparrow$ & 23.74$\pm$0.07 & 26.06$\pm$0.03 & 24.12$\pm$0.06 & 25.81$\pm$0.05 & 25.71$\pm$0.06 & 25.83$\pm$0.04 & \textbf{26.27$\pm$0.02} \\
& CLIP Zero-shot OA $\uparrow$ & 30.40$\pm$0.63 & 80.13$\pm$0.91 & 38.80$\pm$2.14 & 84.00$\pm$0.62 & 73.53$\pm$1.12 & 81.40$\pm$1.11 & \textbf{92.73$\pm$0.64} \\
& FID Score $\downarrow$ & 205.80$\pm$4.01 & 122.45$\pm$2.42 & 118.46$\pm$1.74 & 59.93$\pm$1.51 & 86.65$\pm$2.47 & 53.11$\pm$2.26 & \textbf{41.70$\pm$0.69} \\
& NN-SSIM ($\times$100) $\downarrow$ & \textbf{17.25$\pm$0.14} & 40.14$\pm$0.16 & 30.01$\pm$0.42 & 35.25$\pm$0.17 & 24.72$\pm$0.18 & 33.68$\pm$0.16 & 40.11$\pm$0.11 \\
& IC-Coverage ($\times$100) $\uparrow$ & 72.27$\pm$0.18 & 71.09$\pm$0.10 & 76.43$\pm$0.21 & 76.51$\pm$0.15 & 76.56$\pm$0.15 & 77.66$\pm$0.13 & \textbf{78.29$\pm$0.09} \\
\midrule
\multirow{5}{*}{NWPU} 
& CLIP Score ($\times$100) $\uparrow$ & 22.62$\pm$0.07 & 25.64$\pm$0.02 & 23.02$\pm$0.07 & 25.32$\pm$0.03 & 24.81$\pm$0.05 & 25.37$\pm$0.04 & \textbf{26.03$\pm$0.05} \\
& CLIP Zero-shot OA $\uparrow$ & 17.82$\pm$1.07 & 68.78$\pm$1.21 & 25.00$\pm$0.73 & 70.51$\pm$1.20 & 53.89$\pm$1.18 & 67.91$\pm$1.23 & \textbf{82.18$\pm$0.71} \\
& FID Score $\downarrow$ & 97.39$\pm$0.46 & 18.17$\pm$0.46 & 40.85$\pm$1.03 & 10.72$\pm$0.10 & 25.64$\pm$0.69 & 9.62$\pm$0.41 & \textbf{5.28$\pm$0.06} \\
& NN-SSIM ($\times$100) $\downarrow$ & \textbf{19.55$\pm$0.06} & 37.37$\pm$0.19 & 32.20$\pm$0.23 & 38.25$\pm$0.23 & 23.65$\pm$0.28 & 34.13$\pm$0.24 & 39.86$\pm$0.11 \\
& IC-Coverage ($\times$100) $\uparrow$ & 69.16$\pm$0.13 & 71.77$\pm$0.10 & 76.25$\pm$0.15 & 76.92$\pm$0.13 & 76.89$\pm$0.23 & 78.16$\pm$0.03 & \textbf{78.60$\pm$0.03} \\
\bottomrule
\end{tabular}
}
{\scriptsize
\noindent
\begin{minipage}{\linewidth}
\vspace{.3em}
Note: Best results are shown in \textbf{bold}. For each method, we report the mean and standard deviation over five independently distilled datasets.
\end{minipage}
}
\end{table*}

In this subsection, we further evaluate the distilled samples from five complementary perspectives, including image-text semantic consistency, semantic discriminability, image fidelity, visual proximity to real samples, and intra-class coverage.

\textbf{CLIP Score}.
We evaluate the semantic consistency between the generated samples and their corresponding text prompts ``A satellite image of \texttt{<class name>}'' using the CLIP Score \cite{radford2021learning}, which is defined as the cosine similarity between image and text embeddings extracted by the CLIP-RSICD-v2\footnote{\url{https://huggingface.co/flax-community/clip-rsicd-v2}} model. Higher CLIP Score indicates better image-text semantic alignment.

\textbf{CLIP Zero-shot Classification}.
To assess semantic discriminability, we perform zero-shot classification using the same CLIP-RSICD-v2 model. Specifically, the image embedding is matched against text embeddings of all scene categories, and the predicted category is determined by the highest cosine similarity. We report the OA for CLIP zero-shot classification, where higher values indicate stronger semantic discriminability.

\textbf{FID Score}.
We adopt the Fréchet Inception Distance (FID) \cite{heusel2017gans} to measure the distribution discrepancy between the distilled dataset $\mathcal{D}$ and the real training set $\mathcal{R}$. Specifically, we use the features obtained from the final average pooling layer in the pretrained Inception-v3 model to compute the FID score as
\begin{equation}
{\rm FID}=\|\mu_r-\mu_d\|^2+\mathrm{tr}\left(\Sigma_r+\Sigma_d-2\left(\Sigma_r\Sigma_d\right)^\frac{1}{2}\right),
\label{eq:fid}
\end{equation}
where $(\mu_r,\Sigma_r)$ and $(\mu_d,\Sigma_d)$ denote the mean and covariance of the real and distilled feature distributions, respectively, and $\mathrm{tr}\left(\cdot\right)$ represents the trace linear algebra operation. Lower FID score indicates that the distilled samples better match the distribution of real images.

\textbf{NN-SSIM}.
We compute the average Structural Similarity Index Measure (SSIM)~\cite{wang2004image} between each distilled sample and its top-5 most structurally similar real samples from the same class in the original training set, where the neighbors are ranked by SSIM. This metric reflects the structural proximity between distilled and real training samples, with lower values indicating less visual resemblance to the nearest real samples.

\textbf{IC-Coverage}. We further introduce the intra-class coverage (IC-Coverage) metric to measure how well the distilled samples cover the intra-class visual variation of real data. Specifically, we extract $\ell_2$-normalized features from the global average pooling layer of an ImageNet-pretrained ResNet18. For each class, every real training sample is matched to its most similar distilled sample within the same class, and the resulting maximum cosine similarities are averaged. The final IC-Coverage is obtained by macro-averaging the class-wise scores over all classes. A higher IC-Coverage indicates better coverage of the intra-class diversity of real data.

Table~\ref{tab:metric} summarizes the quality evaluation results based on the above metrics. Overall, DPD achieves the best CLIP Score and CLIP zero-shot classification OA across all three datasets, indicating superior image-text semantic consistency and stronger class-level discriminability. The improvement in CLIP zero-shot OA is particularly significant on more complex datasets. For example, on NWPU, DPD outperforms the strongest baseline DiT-IGD by more than 10 percentage points. In terms of image fidelity, DPD obtains the lowest FID on AID and NWPU, while remaining competitive on UCM. These results demonstrate that the proposed method effectively improves both the realism and semantic reliability of distilled remote sensing samples. For NN-SSIM, MinimaxDiff consistently achieves the lowest values, suggesting lower structural proximity to the original training samples. However, this comes at the cost of semantic consistency and discriminability, as reflected by its inferior CLIP Score and CLIP zero-shot classification OA. In contrast, DPD shows relatively higher structural proximity to real training samples, which is consistent with its design that aims to preserve representative class-specific structures and align the overall visual style of distilled samples with the real data distribution. For IC-Coverage, DPD achieves the best performance on AID and NWPU and remains highly competitive on UCM, where it is only slightly lower than ManifoldGD. These results indicate that DPD effectively covers the intra-class visual variation of real data, demonstrating its strong coverage capability across different datasets.

\begin{figure}[t]
  \centering  
  \includegraphics[width=\linewidth]{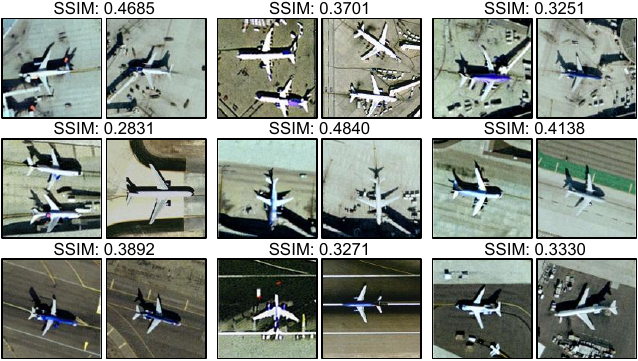}
\caption{Nearest-neighbor visualization of DPD-generated samples on the airplane category of UCM. In each pair, the left image is generated by DPD, and the right image is its SSIM-based nearest neighbor from the real training set.}
\label{fig:ssim}
\end{figure}

Figure~\ref{fig:ssim} further visualizes the SSIM-based nearest-neighbor retrieval results. In each pair, the left image is generated by DPD, while the right image is its SSIM-based nearest neighbor from the real training set. Although the generated samples share similar class-specific structures with their nearest real counterparts, they present distinct spatial layouts or fine-grained visual details. This observation suggests that DPD preserves representative class-level visual patterns without directly reproducing individual training samples.

\begin{figure*}[t]
  \centering  \includegraphics[width=\linewidth]{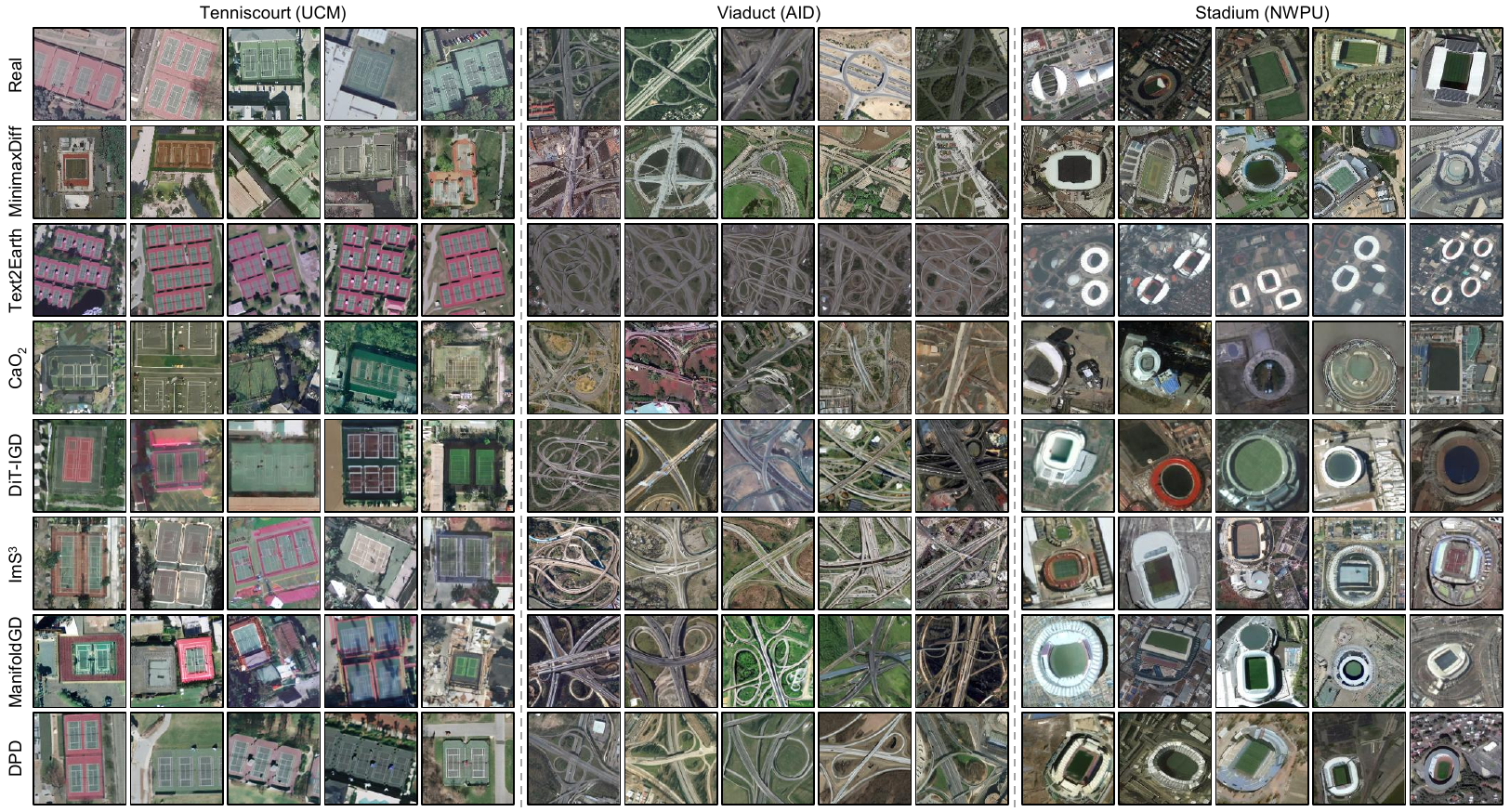}
  \caption{Qualitative comparison of randomly selected real images and distilled samples on the UCM, AID, and NWPU datasets.
}
\label{fig:comparison}
\end{figure*}

\subsection{Qualitative Results}

Figure~\ref{fig:comparison} presents a qualitative comparison between randomly selected real images and distilled images generated by different dataset distillation methods on representative categories from UCM, AID, and NWPU, including \emph{Tenniscourt}, \emph{Viaduct}, and \emph{Stadium}. Overall, the compared methods exhibit varying limitations in preserving realistic remote sensing semantics and maintaining intra-class diversity. For example, Text2Earth tends to generate samples with consistent class-specific patterns, such as repeated tennis courts or stadium-like structures, and often suffers from limited layout diversity and noticeable texture degradation. CaO$_2$ and DiT-IGD can produce more varied layouts in some cases, but their samples occasionally contain blurred object boundaries or weakened structural details. ImS$^3$ and ManifoldGD generally improve visual fidelity, yet some generated samples still show repetitive spatial patterns or incomplete scene structures, especially for complex categories such as \emph{Viaduct} and \emph{Stadium}.

By comparison, the proposed DPD achieves a better balance between visual realism and sample diversity across different datasets and scene categories. For the \emph{Tenniscourt} class in UCM, DPD preserves clear court layouts while presenting diverse colors and surrounding contexts. For the \emph{Viaduct} class in AID, DPD captures complex road intersections and curved overpass structures with relatively realistic spatial organization. For the \emph{Stadium} class in NWPU, DPD generates recognizable stadium shapes with variations in scale and viewpoint. These observations indicate that DPD can effectively preserve key structural and semantic characteristics of real remote sensing scenes while avoiding excessive sample repetition, which is consistent with the quantitative results reported in Section~\ref{quantitative_results}.

\subsection{Ablation Study and Parameter Analysis}
\label{sec:ablation}

\begin{table}[t]
\centering
\caption{Ablation study of the proposed DPD on the UCM, AID, and NWPU datasets with $\text{IPC}=10$.} 
\label{tab:ablation}
\resizebox{\linewidth}{!}{%
\begin{tabular}{ccc|ccc}
\toprule
Direct Gen. & PLD & DLCS & UCM & AID & NWPU \\
\midrule
\checkmark &  &  & 61.10$\pm$2.09 & 53.14$\pm$1.04 & 45.35$\pm$1.44 \\
\checkmark & \checkmark &  & 85.87$\pm$1.94 & 80.82$\pm$0.53 & 72.44$\pm$1.26 \\
\checkmark & \checkmark & \checkmark & \textbf{87.41$\pm$1.28} & \textbf{81.58$\pm$0.75} & \textbf{73.61$\pm$1.08}\\
\bottomrule
\end{tabular}
}
{\scriptsize
\noindent
\begin{minipage}{\linewidth}
\vspace{.3em}
Note: ``Direct Gen.'' denotes direct generation using the pretrained diffusion model without guidance; ``PLD'' denotes prototype-guided latent denoising; ``DLCS'' denotes discriminative latent candidate selection. Best results are shown in \textbf{bold}. Performance is reported in OA (\%) using ResNet18 as the classification network.
\end{minipage}
}
\end{table}

\textbf{Contribution of each component}. Table~\ref{tab:ablation} analyzes the contribution of each component in the proposed DPD framework. The baseline model, which directly generates distilled images using the pretrained diffusion model without additional guidance, achieves limited performance on all three datasets. This indicates that direct generation alone is insufficient to produce highly discriminative distilled samples, even when the diffusion model has been adapted to the target remote sensing dataset. By contrast, introducing prototype-guided latent denoising brings substantial improvements across all datasets, increasing the OA by more than 20 percentage points on UCM, AID, and NWPU. These performance gains demonstrate that visual prototypes provide effective structural and semantic guidance during the denoising process, which helps to preserve representative class-specific patterns and improve the discriminability of the generated samples. Further incorporating discriminative latent candidate selection consistently improves the performance, achieving the best OA values of 87.41\%, 81.58\%, and 73.61\% on UCM, AID, and NWPU, respectively.

\begin{figure}[t]
  \centering  
  \includegraphics[width=\linewidth]{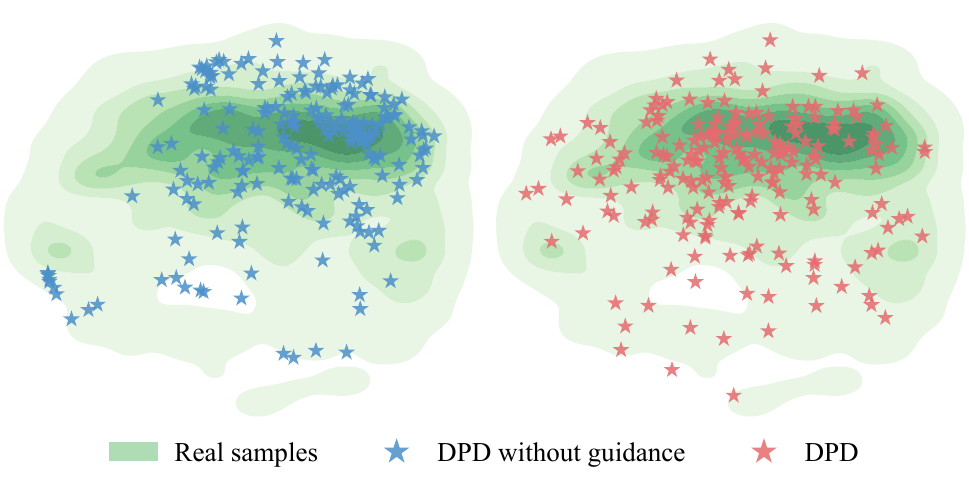}
\caption{
PCA visualization of VAE latent feature distributions on the UCM dataset with $\text{IPC}=10$.
The green density map represents the real training distribution, while blue and red stars denote samples generated by DPD without guidance and DPD, respectively.
}
\label{fig:pca}
\end{figure}

Figure~\ref{fig:pca} visualizes the latent feature distributions of generated samples on UCM. The reference density map is estimated from the real training samples, and the generated samples are projected into the same PCA space. Compared with DPD without guidance, DPD samples are more consistently distributed over the high-density regions of the real-data manifold, while covering a broader latent-space area. This indicates that prototype guidance improves the distributional alignment between generated and real samples.

\begin{table}[t]
\caption{Comparison of different prototype selection strategies for the proposed DPD on UCM, AID, and NWPU with $\text{IPC}=10$.}
\label{tab:prototype_ablation}
\centering
\resizebox{\linewidth}{!}{%
\begin{tabular}{c|cccc}
\toprule
Dataset 
& Random 
& Nearest-to-centroid 
& Direct centroid 
& Proposed \\
\midrule
UCM  
& 83.33$\pm$2.70 
& 86.50$\pm$1.17 
& 84.69$\pm$0.91 
& \textbf{87.41$\pm$1.28} \\
AID  
& 80.01$\pm$0.85 
& 81.43$\pm$1.15 
& 74.28$\pm$1.62 
& \textbf{81.58$\pm$0.75} \\
NWPU 
& 71.80$\pm$0.66 
& 73.19$\pm$0.25 
& 61.87$\pm$2.41 
& \textbf{73.61$\pm$1.08} \\
\bottomrule
\end{tabular}
}
{\scriptsize
\noindent
\begin{minipage}{\linewidth}
\vspace{.3em}
Note: Best results are shown in \textbf{bold}. Performance is reported in OA (\%) using ResNet18 as the classification network.
\end{minipage}
}
\end{table}

\begin{table}[t]
\caption{Parameter analysis of radius scaling factor $\lambda$ on the UCM, AID, and NWPU datasets with $\text{IPC}=10$.}
\label{tab:lambda}
\centering
\resizebox{\linewidth}{!}{%
\begin{tabular}{c|ccccc}
\toprule
$\lambda$ & 0 & 0.1 & 1 & 3 & 10 \\
\midrule
UCM & 86.86$\pm$1.55 & 86.62$\pm$1.84 & \textbf{87.41$\pm$1.28} & 87.05$\pm$1.14 & 85.73$\pm$1.62 \\
AID & 81.34$\pm$1.31 & \textbf{81.87$\pm$1.20} & 81.58$\pm$0.75 & 81.19$\pm$1.07 & 80.68$\pm$0.78 \\
NWPU & 73.12$\pm$0.58 & 74.01$\pm$0.73 & 73.61$\pm$1.08 & \textbf{74.14$\pm$0.78} & 73.89$\pm$1.10 \\
\bottomrule
\end{tabular}}
{\scriptsize
\noindent
\begin{minipage}{\linewidth}
\vspace{.3em}
Note: Best results are shown in \textbf{bold}. Performance is reported in OA (\%) using ResNet18 as the classification network.
\end{minipage}
}
\end{table}

\begin{table}[t]
\caption{Parameter analysis of candidate number $B$ on the UCM, AID, and NWPU datasets with $\text{IPC}=10$.}
\label{tab:B}
\resizebox{\linewidth}{!}{%
\begin{tabular}{c|ccccc}
\toprule
$B$ & 1 & 3 & 5 & 7 & 10 \\
\midrule
UCM & 85.87$\pm$1.94 & 86.23$\pm$1.31 & \textbf{87.41$\pm$1.28} & 86.42$\pm$1.56 & 86.44$\pm$1.66 \\
AID & 80.82$\pm$0.53 & 81.81$\pm$0.87 & 81.58$\pm$0.75 & 81.40$\pm$1.22 & \textbf{82.01$\pm$0.83} \\
NWPU & 72.44$\pm$1.26 & 73.72$\pm$0.85 & 73.61$\pm$1.08 & \textbf{73.84$\pm$1.14} & 73.64$\pm$0.62\\
\bottomrule
\end{tabular}
}
{\scriptsize
\noindent
\begin{minipage}{\linewidth}
\vspace{.3em}
Note: Best results are shown in \textbf{bold}. Performance is reported in OA (\%) using ResNet18 as the classification network.
\end{minipage}
}
\end{table}

\begin{figure*}[t]
\centering
\includegraphics[width=\linewidth]{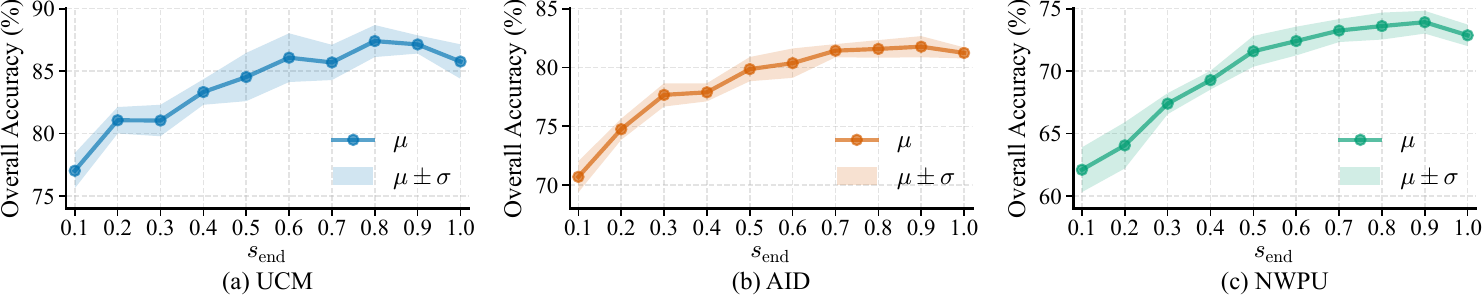}
\caption{Parameter analysis of $s_{\mathrm{end}}$ on the UCM, AID, and NWPU datasets with $\text{IPC}=10$ using ResNet18.}
\label{fig:parameter_analysis}
\end{figure*}

\begin{figure}[t]
  \centering  
  \includegraphics[width=\linewidth]{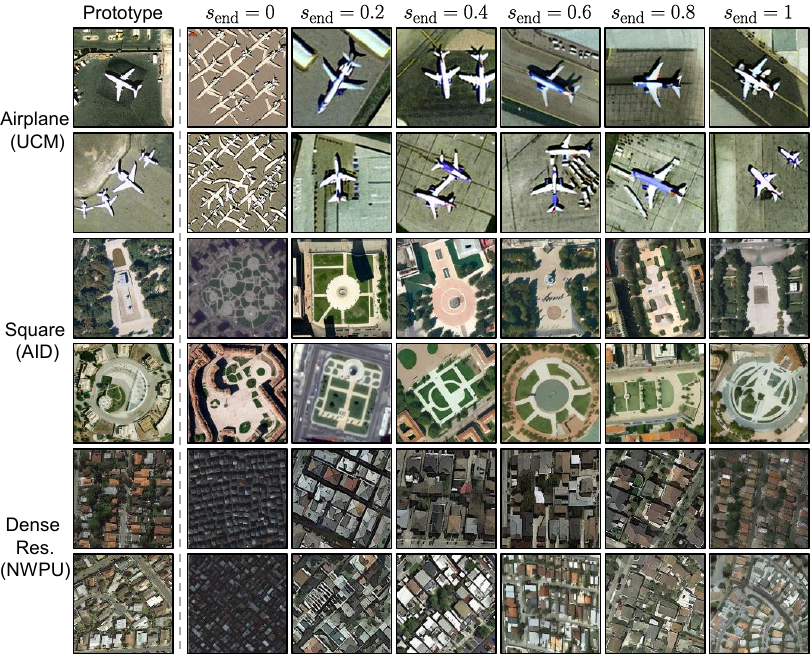}
\caption{Visualization of generated samples with different prototype guidance end ratios $s_{\mathrm{end}}$ on representative classes from UCM, AID, and NWPU. The first column shows the visual prototypes, and the remaining columns show the generated samples as $s_{\mathrm{end}}$ increases from 0 to 1.}
\label{fig:s_end_samples}
\end{figure}

\textbf{Effect of prototype selection strategy}. Table~\ref{tab:prototype_ablation} compares different prototype selection strategies under $\text{IPC}=10$. Here, random, nearest-to-centroid, and direct centroid denote randomly selecting a real sample within each cluster, selecting the real sample closest to the cluster centroid, and directly using the cluster centroid as the prototype, respectively. Compared with random and nearest-to-centroid selection, the results indicate that the proposed strategy, which jointly considers intra-cluster compactness and inter-cluster separability, helps identify more informative prototypes. Moreover, directly using the cluster centroid leads to notably lower performance, particularly on AID and NWPU. This suggests that the centroid itself may not correspond to a representative real sample in the latent space. Overall, the proposed strategy consistently achieves the best performance across all three datasets.

\textbf{Effect of radius $r$}. In our default setting, the radius in the hyperspherical sampling is simply determined by the latent embedding dimension as $r=1/\sqrt{d_{\mathrm{emb}}}$, where $d_{\mathrm{emb}}=4096$. To further investigate the sensitivity to this choice, we introduce a scaling factor $\lambda$ and set $r=\lambda/\sqrt{d_{\mathrm{emb}}}$. Table~\ref{tab:lambda} reports the results with different values of $\lambda$. Overall, the performance is relatively stable over a broad range of radius settings. The best results are obtained at $\lambda=1$, $0.1$, and $3$ on UCM, AID, and NWPU, respectively, while the differences among moderate settings are generally small. In contrast, an excessively large radius ($\lambda=10$) tends to degrade the performance, especially on UCM and AID. This suggests that overly strong deviation from the prototype may weaken the effectiveness of prototype-guided denoising. Given the overall stability across moderate values of $\lambda$, we retain $r=1/\sqrt{d_{\mathrm emb}}$ as a simple and robust default setting without tuning.

\textbf{Effect of candidate number $B$}. Table~\ref{tab:B} analyzes the effect of the candidate number $B$ in discriminative latent candidate selection (Section \ref{sec:discriminative_selection}). When $B=1$, only one sample is generated for each prototype without additional candidate selection. Increasing $B$ generally improves the performance on all three datasets, indicating that generating multiple candidates and selecting the most discriminative ones can effectively enhance the quality of the distilled dataset. For example, compared with $B=1$, the OA increases from 85.87\% to 87.41\% on UCM when $B=5$, from 80.82\% to 82.01\% on AID when $B=10$, and from 72.44\% to 73.84\% on NWPU when $B=7$. Meanwhile, the performance gain tends to saturate as $B$ further increases. Although the best results on AID and NWPU are obtained at $B=10$ and $B=7$, respectively, the performance differences among settings with $B\geq5$ are marginal. This suggests that a moderate number of candidates is sufficient for effective sample selection. Therefore, we set $B=5$ for all datasets to balance performance and generation cost.

\textbf{Effect of end ratio $s_{\mathrm{end}}$}. Figure~\ref{fig:parameter_analysis} analyzes the effect of the prototype guidance end ratio $s_{\mathrm{end}}$ defined in Eq.~\eqref{eq:guidance_window} of Section~\ref{sec:prototype_guided_denoising}. Overall, increasing $s_{\mathrm{end}}$ from a small value consistently improves the performance on all three datasets. This indicates that a longer prototype-guided denoising process enables the generated samples to better exploit the structural and semantic information provided by the visual prototypes, thereby improving the discriminability of the distilled dataset. However, the performance gain gradually saturates when $s_{\mathrm{end}}$ becomes large. On UCM, the best performance is achieved around $s_{\mathrm{end}}=0.8$, while AID and NWPU reach their peak values around $s_{\mathrm{end}}=0.9$. Further increasing the guidance end ratio will reduce the performance, suggesting that excessive prototype guidance could interfere with the final refinement stage of the reverse diffusion process. Therefore, we set $s_{\mathrm{end}}=0.8$ for all datasets in DPD.

Figure~\ref{fig:s_end_samples} further visualizes the effect of $s_{\mathrm{end}}$ on the generated samples. When $s_{\mathrm{end}}=0$, prototype guidance is disabled, and the generated samples may deviate from the prototype semantics or exhibit unstable class-specific structures. As $s_{\mathrm{end}}$ increases, the samples progressively preserve more structural information from the prototypes. However, an overly large $s_{\mathrm{end}}$ may cause the generated samples to inherit excessive prototype-specific details and artifacts, which may reduce generation quality and sample diversity.

\begin{table}[t]
\caption{Comparison of random real-subset, K-means-based coreset selection, and dataset distillation on the UCM dataset.}
\label{tab:coreset}
\centering
\resizebox{\linewidth}{!}{%
\begin{tabular}{c|ccc}
\toprule
IPC (ratio) & Random real subset & K-means coreset & Dataset distillation (DPD) \\
\midrule
5 (10\%)  & 78.95$\pm$1.37 & 83.67$\pm$0.68 & 80.10$\pm$0.70 \\
10 (20\%) & 90.36$\pm$1.27 & 92.37$\pm$0.45 & 87.41$\pm$1.28 \\
15 (30\%) & 93.52$\pm$0.93 & 94.23$\pm$0.50 & 88.53$\pm$0.56 \\
20 (40\%) & 95.47$\pm$0.47 & 96.25$\pm$0.45 & 90.02$\pm$0.89 \\
\bottomrule
\end{tabular}
}
{\scriptsize
\noindent
\begin{minipage}{\linewidth}
\vspace{.3em}
Note: Performance is reported in OA (\%) using ResNet18 as the classification network.
\end{minipage}
}
\end{table}

\textbf{Comparison with data-reduction strategies}. To further examine the practical trade-off between dataset distillation and direct real-data reduction, we compare DPD with two simple strategies on UCM under the same IPC budget: random real-subset selection and K-means-based coreset selection. Random selection samples the required number of real training images from each class, whereas K-means-based coreset selection clusters the VAE latent representations within each class and retains the real sample closest to each cluster centroid. As shown in Table~\ref{tab:coreset}, DPD outperforms random subset selection at $\text{IPC}=5$, whereas directly retaining real samples becomes increasingly advantageous as the IPC budget grows, with K-means-based coreset selection consistently achieving the highest accuracy. This trend is particularly evident at $\text{IPC}=20$, where 40\% of the original training samples are retained. These results show that representative real-sample selection provides a strong alternative when a substantial fraction of the training set can be preserved. In contrast, dataset distillation constructs a synthetic surrogate set without directly retaining the original training samples. The observed accuracy gap also indicates that improving the discriminative representativeness of distilled samples, particularly in the presence of the large intra-class variability and complex visual patterns of remote sensing data, remains an important direction for future work.

\section{Conclusion}
\label{conclusion}
In this paper, we present the first systematic study of dataset distillation for remote sensing image classification and demonstrate that large-scale remote sensing datasets can be effectively condensed into compact, fully synthetic alternatives. Specifically, we propose a discriminative prototype-guided diffusion (DPD) framework that exploits representative latent prototypes as semantic anchors to guide the reverse denoising trajectory. Furthermore, we introduce a discriminative latent candidate selection strategy that ranks multiple generated candidates with a latent classifier and selects samples with higher logit margins to enhance the discriminative quality of the distilled dataset. Extensive experiments on three high-resolution remote sensing scene classification benchmarks validate the effectiveness of the proposed method in supporting downstream model training.

Since this work mainly focuses on dataset distillation for image-level scene classification, extending existing distillation frameworks to more fine-grained remote sensing interpretation tasks, such as object detection and land-cover mapping, still remains an open challenge. We will explore it in our future research.

\bibliographystyle{IEEEtran}

\bibliography{dpd}

\end{document}